\documentclass[sigconf]{acmart}
\usepackage{bbding}
\AtBeginDocument{%
  }

\copyrightyear{2025}
\acmYear{2025}
\setcopyright{acmlicensed}
\acmConference[MM '25]{Proceedings of the 33rd ACM International Conference on Multimedia}{October 27--31, 2025}{Dublin, Ireland}
\acmBooktitle{Proceedings of the 33rd ACM International Conference on Multimedia (MM '25), October 27--31, 2025, Dublin, Ireland}
\acmDOI{10.1145/3746027.3755504}
\acmISBN{979-8-4007-2035-2/2025/10}

\begin{document}

\title{Cross Paradigm Representation and Alignment Transformer for Image Deraining}


\author{Shun Zou$^{\dagger}$}
\affiliation{%
  \institution{Nanjing Agricultural University}
  \city{Nanjing}
  \country{China}}
\affiliation{%
  \institution{Soochow University}
  \city{Suzhou}
  \country{China}}
\email{zs@stu.njau.edu.cn}

\author{Yi Zou$^{\dagger}$}
\affiliation{%
  \institution{Xiangtan University}
  \city{Xiangtan}
  \country{China}
}
\email{202205570112@smail.xtu.edu.cn}

\author{Juncheng Li}
\affiliation{%
 \institution{Shanghai University}
 \city{Shanghai}
 \country{China}}
 \email{cvjunchengli@gmail.com}

\author{Guangwei Gao$^{\ast}$}
\affiliation{%
  \institution{Nanjing University of Posts and Telecommunications}
  \city{Nanjing}
  \country{China}}
\affiliation{%
  \institution{Soochow University}
  \city{Suzhou}
  \country{China}}
\email{csggao@gmail.com}

\author{Guo-jun Qi}
\affiliation{%
  \institution{Westlake University}
  \city{Hangzhou}
  \country{China}}
\email{guojunq@gmail.com}

\thanks{$^{\dagger}$These authors contributed equally to this work, $^{\ast}$Corresponding author.}

\renewcommand{\shortauthors}{Shun Zou, Yi Zou, Juncheng Li, Guangwei Gao, and Guo-jun Qi}
\begin{abstract}
Transformer-based networks have achieved strong performance in low-level vision tasks like image deraining by utilizing spatial or channel-wise self-attention. However, irregular rain patterns and complex geometric overlaps challenge single-paradigm architectures, necessitating a unified framework to integrate complementary global-local and spatial-channel representations.
To address this, we propose a novel Cross Paradigm Representation and Alignment Transformer (CPRAformer). Its core idea is the hierarchical representation and alignment, leveraging the strengths of both paradigms (spatial-channel and global-local) to aid image reconstruction. It bridges the gap within and between paradigms, aligning and coordinating them to enable deep interaction and fusion of features.
Specifically, we use two types of self-attention in the Transformer blocks: sparse prompt channel self-attention (SPC-SA) and spatial pixel refinement self-attention (SPR-SA). SPC-SA enhances global channel dependencies through dynamic sparsity, while SPR-SA focuses on spatial rain distribution and fine-grained texture recovery. To address the feature misalignment and knowledge differences between them, we introduce the Adaptive Alignment Frequency Module (AAFM), which aligns and interacts with features in a two-stage progressive manner, enabling adaptive guidance and complementarity. This reduces the information gap within and between paradigms.
Through this unified cross-paradigm dynamic interaction framework, we achieve the extraction of the most valuable interactive fusion information from the two paradigms.
Extensive experiments demonstrate that our model achieves state-of-the-art performance on eight benchmark datasets and further validates CPRAformer's robustness in other image restoration tasks and downstream applications.

\end{abstract}





\maketitle
\begin{figure}[ht]
	\centering
	\includegraphics[width=\linewidth]{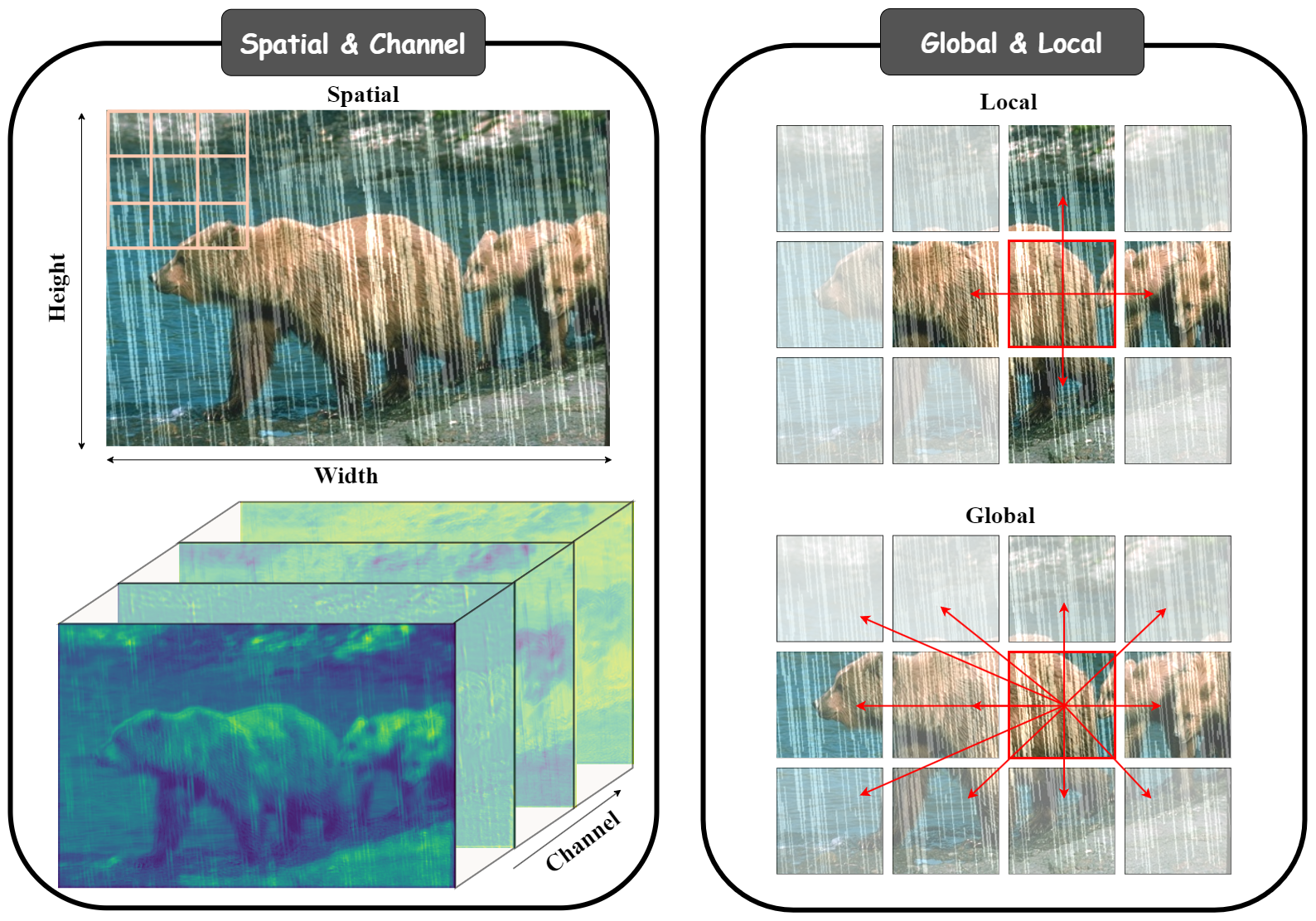} 
	\caption{Feature patterns obtained from four perspectives are distinct, two deraining paradigms offers unique advantages. Recent deraining research mainly focuses on spatial-channel or global-local paradigms, lacking a framework that effectively integrates these two paradigms.}
	\label{paradigms}
	\vspace{-6mm}
\end{figure}

\section{Introduction}
Single Image Deraining (SID) is a traditional low-level vision task that aims to restore a clear, high-quality image from a given rainy image. As it plays a critical role in downstream tasks across various fields, including video surveillance, autonomous driving, and medical imaging, it has garnered increasing attention from both academia and industry. Due to its ill-posed nature, early methods typically applied various priors based on the statistical characteristics of rain streaks and clean images \cite{li2016rain,zhang2017convolutional}. However, in complex and diverse rainy scenarios, such priors do not always hold.

Recently, many studies have proposed convolutional neural network (CNN)-based methods to address this challenge \cite{9857285,Kui_2020_CVPR,ren2019progressive,Zamir2021MPRNet,yi2021structure,li2018recurrent}. However, due to the limited receptive field of convolution operators, long-range spatial modeling is hindered, which restricts model performance. Fortunately, inspired by the success of Transformers in natural language processing and advanced vision tasks \cite{vaswani2017attention,dosovitskiy2020vit,tu2022maxvit,wang2021pyramid}, researchers have developed Transformer-based architectures for the SID task \cite{Wang_2022_CVPR,Chen_2023_CVPR,valanarasu2022transweather}. Leveraging the self-attention mechanism, Transformer-based methods can establish global dependencies, alleviating the limitations of CNN-based approaches and demonstrating superior deraining performance. Recognizing the potential of Transformers, some researchers have explored their effective application in SID tasks from different perspectives. 
In terms of spatial modeling, some methods use non-overlapping spatial windows to capture more global dependencies, enhancing spatial pixel modeling \cite{Wang_2022_CVPR}. Regarding channels, "transpose" attention has been proposed \cite{Zamir2021Restormer}, where self-attention is computed along the channel dimension instead of the spatial dimension. These methods, with strong feature extraction and modeling capabilities in their respective dimensions, have achieved remarkable results. Intuitively, extracting spatial features and capturing channel context information both play crucial roles in enhancing Transformer performance in image restoration \cite{chen2023dual}. Additionally, due to the intricate intertwining of rain streaks and the rain-free background, both global and local features are essential for the challenging SID task. However, the self-attention mechanism in Transformers does not fully leverage the local invariance of CNNs. To address this, some researchers have attempted to combine CNNs with Transformers \cite{chen2023hybrid,Chen_2023_CVPR}, inheriting CNNs' advantage in local modeling and Transformers' strength in capturing long-range dependencies.

However, two key questions naturally arise: (1) How can we simultaneously leverage information from all paradigms? As shown in Fig. \ref{paradigms}, representations from different perspectives undoubtedly play a critical role in SID task performance; (2) How can we effectively align and aggregate the two feature types within each paradigm? The distinct feature models across paradigms make simple summation or concatenation prone to information loss, failing to significantly boost performance. To address these issues, we propose a novel hybrid architecture: the Cross Paradigm Representation and Alignment Transformer (CPRAformer). Its core idea is to establish a cross-paradigm representation learning framework through dimensional consistency (spatial-channel perspective) and multi-perspective integration (global-local perspective), along with alignment and hierarchical fusion of corresponding features in each paradigm.
Specifically, it comprises two carefully designed components: Cross-Paradigm Interaction and Alignment Self-Attention (CPIA-SA) and Multi-Scale Flow Gating Network (MSGN). 

In CPIA-SA, we introduce two types of self-attention: Sparse Prompt Channel Self-Attention (SPC-SA) and Spatial Pixel Refinement Self-Attention (SPR-SA). SPC-SA computes attention along the channel dimension and dynamically filters attention values in the dense attention matrix using prompt information. This allows the network to exploit sparsity, retaining the most valuable attention information while minimizing excessive noise interactions that could degrade image restoration quality, thus effectively extracting global-channel information. Conversely, SPR-SA utilizes an efficient CNN-based architecture to approximate self-attention, enabling the effective modeling of local fine-grained features and the relationships between neighboring spatial pixels, fully leveraging local spatial characteristics.
Additionally, these two self-attention mechanisms are complementary. SPC-SA provides global information between features for SPR-SA, thereby expanding the receptive field of pixels. Meanwhile, SPR-SA enhances the spatial representation of each feature map, which aids in modeling channel context.

At the same time, to further promote alignment and interaction within each paradigm, we propose a two-stage progressive fusion strategy called the Adaptive Alignment Frequency Module (AAFM). Using adaptive weighting, it aligns corresponding branches and enhances interactions between frequency spectra to aggregate and strengthen internal feature information.
Moreover, another key component of the Transformer module is the feed-forward network (FFN) \cite{dosovitskiy2020vit}, which typically extracts features through fully connected layers but often overlooks the critical multi-scale information needed for SID tasks \cite{chen2024rethinking,Chen_2023_CVPR}. To address this limitation, we introduce the Multi-Scale Flow Gating Network (MSGN). Utilizing a gating mechanism, MSGN incorporates multi-scale representation learning, providing additional nonlinear information to the FFN and enhancing its capability to capture essential features.

In summary, through the design described above, our CPRAformer achieves cross-paradigm feature pattern learning and information alignment, thereby enabling robust feature representation.
Our main contributions are summarized as follows:
\begin{itemize}
    \item 
    We propose a hybrid model for SID, CPRAformer, which integrates the advantages of spatial-channel and global-local paradigms. Through cross-paradigm dynamic interaction, it aligns and adaptively fuses feature patterns between the two paradigms.
    %
    \item We utilize both SPC-SA and SPR-SA to extract features across spatial and channel dimensions, effectively modeling global dependencies while capturing local details for complementary feature integration. 

    \item To bridge the feature gap between paradigms, we develop the AAFM, which first performs feature alignment through adaptive weighting and then achieves feature fusion via frequency-domain interaction. This facilitates better coordination and deep interaction between different types of information. In addition, the MSGN is also included to learn scale-aware spatial features.
    %
\end{itemize}

\begin{figure*}[t]
	\centering
	\includegraphics[width=\linewidth]{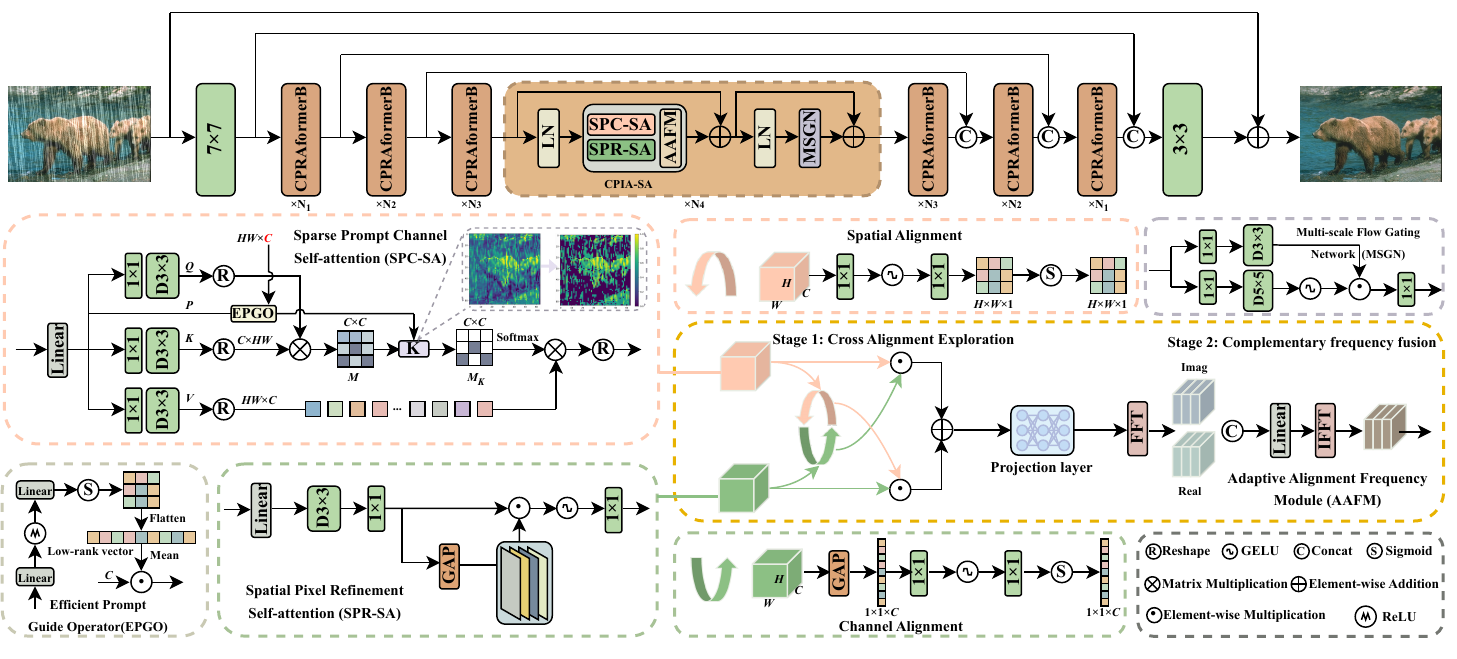} 
	\vspace{-8mm}
	\caption{The overall architecture of our proposed CPRAformer.}
	\label{overview}
	\vspace{-4mm}
\end{figure*}

\section{Related works}
\subsection{Single Image Deraining}
Traditional rain removal methods often rely on handcrafted priors \cite{li2016rain, zhang2017convolutional, gu2017joint, peng2023rain2avoid, kang2011automatic}, which are subjective and unable to adapt to complex rainy scenes. To address this issue, many researchers have developed CNN-based methods \cite{9857285, Kui_2020_CVPR, ren2019progressive, Zamir2021MPRNet, yi2021structure, li2018recurrent, yang2020single, yu2022towards, wang2020dcsfn} for image deraining, achieving promising results. However, convolutions struggle to capture long-range dependencies in both spatial and channel dimensions. Inspired by the success of Transformers in advanced vision tasks \cite{vaswani2017attention, dosovitskiy2020vit, tu2022maxvit, wang2021pyramid}, they have also been applied to image deraining \cite{Wang_2022_CVPR, Chen_2023_CVPR, valanarasu2022transweather, wang2021pyramid, wang2023hpcnet, ren2023semi, liang2022drt, xiao2022image, wang2023promptrestorer}. Transformers, as a new network backbone, show significant improvements over CNN-based methods due to their excellent global context awareness enabled by self-attention. However, a limitation of self-attention is that tokens with low attention values or irrelevant tokens can interfere with the dense attention matrix, potentially harming output features \cite{Chen_2023_CVPR, xiao2024ttst,su2024high}. As shown in Fig. \ref{attention}, to overcome this, we propose an adaptive sparse attention mechanism, which dynamically adjusts the sparse range using a learnable operator. This approach maximizes network sparsity and reduces excessive interference from irrelevant noise in naive self-attention.

\vspace{-3mm}
\subsection{Feature Aggregation}
{\flushleft\textbf{Global \& Local.}}
To leverage the advantage of CNNs in extracting local features and Transformers in global modeling, many hybrid models have been proposed. For example, ELF \cite{magic_elf} was the first to unify these architectures into a lightweight deraining model based on association learning. Inspired by progressive learning, HCT-FFN \cite{chen2023hybrid} introduced a new staged hybrid deraining network. SMFANet \cite{zheng2024smfanet} uses adaptive feature aggregation to synergize local and non-local feature interactions. Dual-former \cite{chen2024dual} combines the global modeling power of self-attention with the local capability of convolutions in a unified architecture. 
It uses a hybrid Transformer block to model long-range spatial dependencies and handle uneven channel distributions.
\vspace{-2mm}
{\flushleft\textbf{Spatial \& Channel.}}
In CNNs, researchers apply attention mechanisms along the spatial and channel dimensions to enhance feature representation, as demonstrated by models like RESCAN \cite{li2018recurrent}, MSPFN \cite{Kui_2020_CVPR}, and MPRNet \cite{Zamir2021MPRNet}. In Transformer-based approaches, spatial self-attention is primarily used to capture long-range dependencies between pixels; for instance, IDT \cite{9798773} uses dual Transformers with window and spatial attention for deraining. Additionally, some studies integrate channel self-attention in Transformers to combine spatial and channel information. Notably, Restormer \cite{Zamir2021Restormer} designs an efficient Transformer model by estimating self-attention along the channel dimension, achieving significant performance gains, while DRSFormer \cite{Chen_2023_CVPR} proposes a sparse Transformer along the channel dimension to fully exploit the most informative features for deraining. DAT \cite{chen2023dual} alternates between spatial and channel self-attention mechanisms across consecutive Transformer blocks, aggregating features from the spatial and channel dimensions both across and within blocks.
\vspace{-2mm}
{\flushleft\textbf{Motivation.}}
Extensive research on both paradigms, namely global-local as well as spatial-channel, demonstrates the critical role of feature representation in complex rainy scenes. However, no study has simultaneously considered both paradigms (see Fig. \ref{paradigms}). Therefore, we introduce two types of self-attention mechanisms into image deraining for comprehensive information exploration and representation. Additionally, we design AAFM to dynamically achieve both inter-paradigm and intra-paradigm feature aggregation, iteratively refining feature coherence across scales and dimensions.

\section{Method}

\subsection{Overall Pipeline}
The overall pipeline of our proposed CPRAformer is illustrated in Fig. \ref{overview}, which adopts an encoder-decoder architecture with skip connections. Specifically, given an input rain image \( I_{\text{rain}} \in \mathbb{R}^{H \times W \times 3} \), we first apply a \( 7 \times 7 \) convolution to obtain the low-level feature embedding \( F_0 \in \mathbb{R}^{H \times W \times C} \), where \( H \), \( W \), and \( C \) represent the height, width, and channels, respectively. The low-level feature embedding is then fed into the backbone network, which consists of a 4-level encoder-decoder structure.
In the encoder stage, the resolution of the high-resolution input is reduced by a factor of 2 while increasing the number of channels, whereas in the decoder stage, the process is reversed. Each encoder and decoder is composed of \( N_i \) ( \( i = 1, 2, 3, 4 \)) stacked Cross Paradigm Representation and Alignment Transformer Block (CPRAformerBs). Within each CPRAformerB, we develop the CPIA-SA to extract and aggregate features from both spatial-channel and global-local paradigms simultaneously. Additionally, we incorporate MSGN within each CPRAformerB, which leverages an elegant gating mechanism to extract multi-scale information, aiding in the image-deraining process.
%

As shown in Fig. \ref{overview}, CPIA-SA is composed of three main components: Sparse Prompt Channel Self-Attention (SPC-SA), Spatial Pixel Refinement Self-Attention (SPR-SA), and Frequency Adaptive Interaction Module (AAFM). 
SPC-SA, using the Efficient Prompt Guide Operator (EPGO), effectively explores the sparsity of the neural network, adaptively retaining the most valuable attention values. This enables efficient extraction of rain-degraded features that handle spatial variations while modeling global channel context. In contrast, SPR-SA focuses on modeling the spatial background, enhancing the spatial pixel representation of each feature map, and facilitating accurate background restoration through fine-grained local features.
To bridge the gap between SPC-SA and SPR-SA in terms of spatial-channel and global-local knowledge, and to fully integrate features within both paradigms, we introduce AAFM. AAFM utilizes a dual-stage progressive strategy to alignment and fuse paradigm-specific features comprehensively.

\subsection{Sparse Prompt Channel Self-Attention}
\label{SPR-SA}
As shown in Fig. \ref{overview}, the self-attention mechanism in SPC-SA operates along the channel dimension. Specifically, given the input embedding feature \( F \in \mathbb{R}^{H \times W \times C} \), point-wise convolution (PWConv) and 3×3 depth-wise convolution (DWConv) are applied to \( F \) to aggregate cross-pixel channel information, generating the matrices for query \( Q \), key \( K \), and value \( V \). 
Next, we perform a dot-product operation on the reshaped \( Q \) and \( K \) to generate a dense attention matrix \( M \in \mathbb{R}^{C \times C} \). However, we observe that the tokens in the keys are not always relevant to those in the queries, and the self-attention values estimated using irrelevant tokens introduce noise interactions and information redundancy, affecting the quality of image recovery. 
To address this, we introduce a Top-k mechanism, which differs from traditional self-attention mechanisms, to filter the information in the attention matrix, retaining the most significant attention values and avoiding noise that can lead to artifacts in the deraining process. For example, with \( k = 4/5 \), we only retain the top 80\% of attention scores, while the remaining elements are masked to zero.
Notably, we also develop a novel learnable operator: the Efficient Prompt Guide Operator (EPGO), which dynamically generates prompt information based on the input, guiding the \( K \) values to achieve adaptive modulation and facilitating a dynamic selection process in the attention matrix. As shown in Fig. \ref{attention}, unlike previous studies \cite{Zamir2021Restormer,Chen_2023_CVPR}, we propose a novel dynamic sparse mechanism that fully exploits the sparsity of neural networks. Formally, the above process is expressed as follows:
        
\begin{figure}[t]
	\centering
	\includegraphics[width=\linewidth]{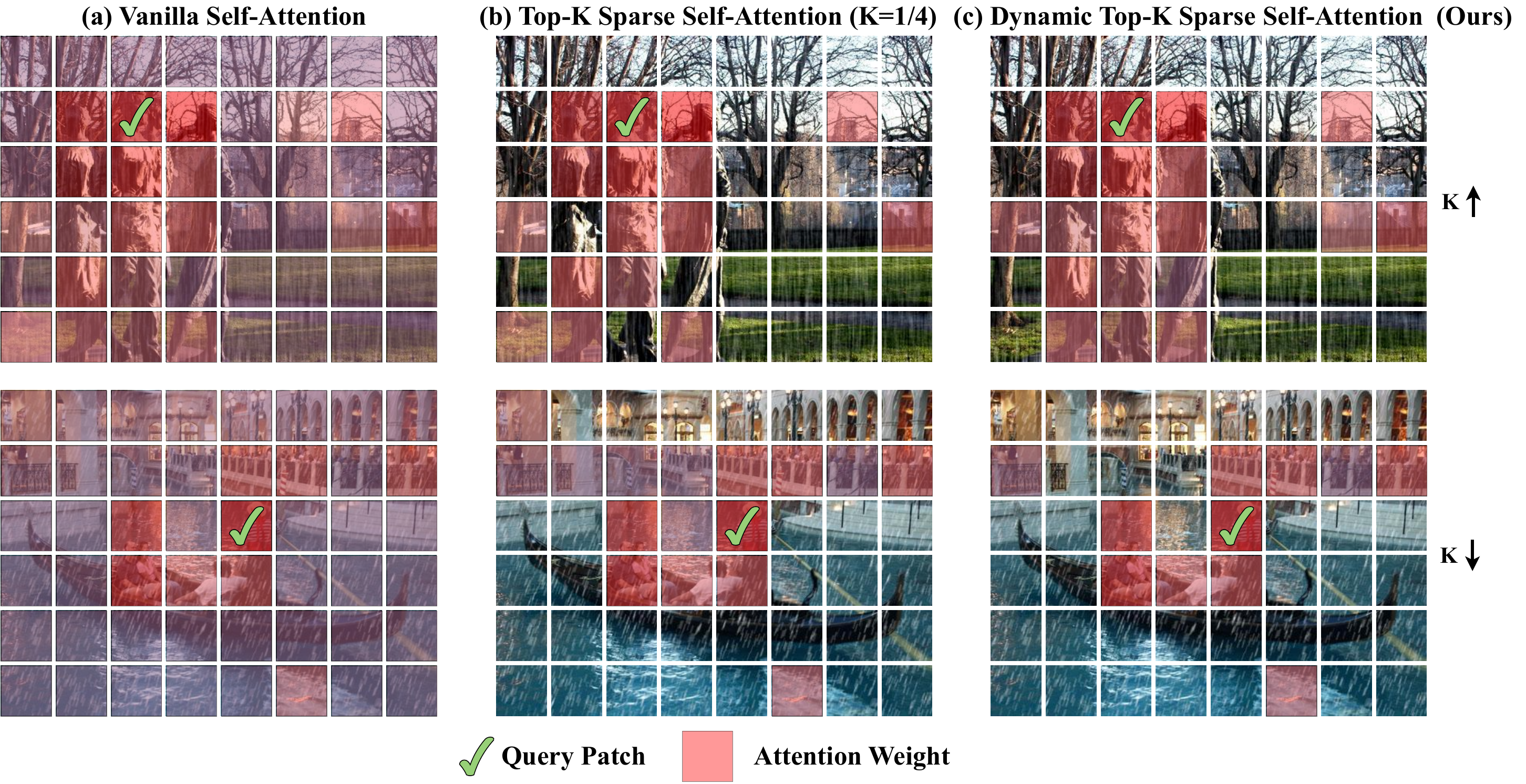} 
    \vspace{-7mm}
	\caption{Comparison of different self-attention mechanisms. (a) The naive self-attention mechanism \cite{Zamir2021Restormer} computes and retains all tokens. (b) The Top-K sparse attention mechanism \cite{Chen_2023_CVPR} sets a fixed K value (here, K is set to 1/4) and retains only the top K\% tokens with the highest attention values while setting the remaining tokens to zero. (c) Our dynamic Top-K sparse attention mechanism adaptively modulates the K value based on input features. For instance, compared to the fixed K in (b), the K value increases in the upper image and decreases in the lower image to adapt to different images.}
	\label{attention}
	\vspace{-4mm}
\end{figure}

\vspace{-1mm}
\begin{equation}
    \Omega_i^{(k)}=\arg\max_{S\subset\{1,\ldots,N\}}\sum_{j\in S}M_{ij},
    \vspace{-1mm}
\end{equation}
\begin{equation}
    \Omega^{(k)}=\left\{\Omega_1^{(k)},\Omega_2^{(k)},\ldots,\Omega_N^{(k)}\right\},
    \vspace{-1mm}
\end{equation}
\begin{equation}
    [M_{k}]_{ij}=
    \begin{cases}
    1, & \mathrm{if~}j\in\Omega_{i}^{(k)}  \Longleftrightarrow\quad M_{ij}\in\mathrm{Top}_{k}(M_{i,:}), \\ 
    0, & \text{otherwise.} 
    \end{cases}
\vspace{-1mm}
\end{equation}
Here, \( \Omega_i^{(k)} \) represents the index set of the top-k elements in the i-th row, that is, the column indices of the k most important elements in the i-th row of matrix \( M \). Formally, the process of SPC-SA is expressed as
\vspace{-2mm}
\begin{equation}
    SPC\text{-}SA=Softmax(T_{k}(\frac{QK^{T}}{\sqrt{d} })) V,
    \vspace{-2mm}
\end{equation}
where \( T_k(\cdot) \) represents the Top-k selection operation after modulation by the Efficient Prompt Guide Operator., and \( \sqrt{d} \) represents an optional temperature coefficient used to control the magnitude of the dot product between \( Q \) and \( K \) before applying softmax. Similar to most previous works \cite{dosovitskiy2020vit}, we employ a multi-head strategy, concatenating all outputs of the multi-head attention and then obtaining the final result through a linear projection.

\begin{table*}
\begin{center}
\caption{Comparison of quantitative results on five datasets. Bold and underlined indicate the best and second-best results.}
\label{table:deraining}
\vspace{-2mm}
\setlength{\tabcolsep}{8pt}
\scalebox{0.72}{
\begin{tabular}{l c c c c c c c c c c c || c c}
\toprule[0.15em]
  & & \multicolumn{2}{c}{\textbf{Test100}~\cite{zhang2019image}}&\multicolumn{2}{c}{\textbf{Rain100H}~\cite{yang2017deep}}&\multicolumn{2}{c}{\textbf{Rain100L}~\cite{yang2017deep}}&\multicolumn{2}{c}{\textbf{Test2800}~\cite{fu2017removing}}&\multicolumn{2}{c||}{\textbf{Test1200}~\cite{zhang2018density}}&\multicolumn{2}{c}{\textbf{Average}}\\
 \textbf{Method} & \textbf{Year} &
 PSNR~$\textcolor{black}{\uparrow}$ & SSIM~$\textcolor{black}{\uparrow}$ & PSNR~$\textcolor{black}{\uparrow}$ & SSIM~$\textcolor{black}{\uparrow}$ & PSNR~$\textcolor{black}{\uparrow}$ & SSIM~$\textcolor{black}{\uparrow}$ & PSNR~$\textcolor{black}{\uparrow}$ & SSIM~$\textcolor{black}{\uparrow}$ & PSNR~$\textcolor{black}{\uparrow}$ & SSIM~$\textcolor{black}{\uparrow}$ & PSNR~$\textcolor{black}{\uparrow}$ & SSIM~$\textcolor{black}{\uparrow}$\\
\midrule[0.15em]
RESCAN~\cite{li2018recurrent} & ECCV2018 & 21.59  & 0.726  & 18.01  & 0.467  & 24.15  & 0.791  & 24.50  & 0.765  & 24.40  & 0.759  & 22.53 & 0.702    \\

PReNet~\cite{ren2019progressive} & CVPR2019 & 23.17  & 0.752  & 17.63  & 0.487  & 27.76  & 0.876  & 27.20  & 0.825  & 26.05  & 0.792  & 24.36 & 0.746    \\

SPDNet~\cite{yi2021structure} & ICCV2021 & 24.25  & 0.848  & 25.87  & 0.809  & 28.63  & 0.880  & 31.05  & 0.904  & 30.42  & 0.893  & 28.04 & 0.867    \\

PCNet~\cite{jiang2021rain} & TIP2021 & 23.29  & 0.762  & 20.83  & 0.563  & 26.64  & 0.817  & 27.10  & 0.818  & 26.53  & 0.791  & 24.88 & 0.750    \\

MPRNet~\cite{Zamir2021MPRNet} & CVPR2021 & 25.66  & 0.859  & 28.23  & 0.850  & 31.94  & 0.930  & 32.14  & 0.925  & 31.32  & 0.901  & 29.86 & 0.893    \\

HINet~\cite{Chen_2021_CVPR} & CVPRW2021 & 23.21  & 0.767  & 20.85  & 0.598  & 27.03  & 0.842  & 28.36  & 0.843  & 27.77  & 0.821  & 25.44 & 0.774    \\

DANet~\cite{jiangDANet2022} & IJCAI2022 & 23.96  & 0.839  & 23.00  & 0.791  & 29.51  & 0.906  & 30.32  & 0.903  & 29.99  & 0.888  & 27.36 & 0.865    \\

Uformer~\cite{Wang_2022_CVPR} & CVPR2022 & 23.87  & 0.815  & 22.43  & 0.700  & 28.39  & 0.883  & 29.71  & 0.886  & 28.65  & 0.856  & 26.61 & 0.828    \\

ALformer~\cite{magic_elf} & ACMMM2022 & 24.41  & 0.844  & 25.10  & 0.807  & 29.39  & 0.903  & 31.36  & 0.916  & 30.40  & 0.897  & 28.13 & 0.874    \\

NAFNet~\cite{chen2022simple} & ECCV2022 & 25.75  & 0.845  & 26.76  & 0.813  & 31.27  & 0.925  & 31.71  & 0.918  & 30.62  & 0.892  & 29.22 & 0.879    \\

MIRNetV2~\cite{Zamir2022MIRNetv2} & TPAMI2022 & 25.76  & 0.867  & 28.05  & 0.846  & 32.53  & 0.935  & 32.33  & 0.925  & 32.38  & 0.915  & 30.21 & 0.897    \\

MFDNet~\cite{10348527} & TIP2023 & 25.90  & 0.870  & 27.06  & 0.850  & 32.76  & 0.944  & 31.92  & 0.925  & 31.15  & 0.909  & 29.76 & 0.899    \\

HCT-FFN~\cite{chen2023hybrid} & AAAI2023 & 24.86  & 0.847  & 26.70  & 0.819  & 29.94  & 0.906  & 31.46  & 0.915  & 31.23  & 0.901  & 28.84 & 0.878    \\

DRSformer~\cite{Chen_2023_CVPR} & CVPR2023 & 27.86  & 0.885  & 28.16  & 0.864  & 34.79  & 0.954  & 32.80  & \underline{0.931}  & 30.99  & 0.906  & 30.92 & 0.908    \\

ChaIR~\cite{cui2023exploring} & KBS2023 & 28.19  & 0.879  & 28.69  & 0.862  & 34.52  & 0.953  & \underline{32.85}  & \underline{0.931}  & 31.30  & 0.903  & 31.11 & 0.906    \\

IRNeXT~\cite{cui_ir} & ICML2023 & 25.80  & 0.860  & 27.22  & 0.833  & 31.65  & 0.931  & 30.53  & 0.917  & 29.02  & 0.898  & 28.85 & 0.888    \\

OKNet~\cite{cui2024omni} & AAAI2024 & 25.43  & 0.858  & 24.01  & 0.804  & 31.19  & 0.928  & 29.32  & 0.911  & 27.56  & 0.886  & 27.50 & 0.877    \\

AST~\cite{zhou2024AST} & CVPR2024 & 26.07  & 0.859  & 27.40  & 0.833  & 32.03  & 0.932  & 31.65  & 0.921  & 30.69  & 0.897  & 29.57 & 0.889    \\

SFHformer~\cite{SFHformer} & ECCV2024 & 25.67  & 0.856  & 27.25  & 0.832  & 32.97  & 0.944  & 32.27  & 0.925  & 31.50  & 0.904  & 29.94 & 0.892    \\

Nerd-rain~\cite{NeRD-Rain} & CVPR2024 & 27.16  & 0.869  & 28.07  & 0.838  & 33.72  & 0.949  & 32.63  & 0.927  & 30.45  & 0.890  & 30.41 & 0.895    \\

MSDT~\cite{chen2024rethinking} & AAAI2024 & 27.79  & 0.878  & 29.05  & 0.856  & 34.75  & 0.955  & 32.68  & 0.930  & \textbf{32.12}  & \textbf{0.917}  & 31.28 & 0.907    \\

FSNet~\cite{FSNet} & TPAMI2024 & 27.95  & 0.884  & 28.70  & 0.860  & 34.10  & 0.952  & 32.68  & \underline{0.931}  & 31.26  & 0.910  & 30.94 & 0.908    \\

AdaIR~\cite{cui2025adair} & ICLR2025 & \underline{28.64} & \underline{0.889} & \underline{29.48} & \underline{0.871} & \underline{35.84} & \underline{0.962} & 32.70 & 0.930 & 30.58 & 0.907 & \underline{31.45} &  \underline{0.912}\\

\midrule

\textbf{CPRAformer (Ours)} & -- & \textbf{29.65} & \textbf{0.895} & \textbf{29.68} & \textbf{0.875} & \textbf{35.98} & \textbf{0.964} & \textbf{33.00} & \textbf{0.933} & \underline{31.52} & \underline{0.913} & \textbf{31.97} & \textbf{0.916} \\

\bottomrule[0.15em]
\end{tabular}}

\end{center}
\vspace{-4mm}
\end{table*}
\begin{table}
\begin{center}
\caption{Quantitative evaluations on Raindrop dataset \cite{Qian_2018_CVPR}. Bold and underlined indicate the best and second-best results.}
\label{table:raindrop}
\vspace{-2mm}
\setlength{\tabcolsep}{8pt}
\scalebox{0.68}{
\begin{tabular}{l c c c c c c c}
\toprule[0.15em]
  & & \multicolumn{2}{c}{\textbf{Raindrop-A}~\cite{Qian_2018_CVPR}}&\multicolumn{2}{c}{\textbf{Raindrop-B}~\cite{Qian_2018_CVPR}}\\
 \textbf{Method} & \textbf{Year} &
 PSNR~$\textcolor{black}{\uparrow}$ & SSIM~$\textcolor{black}{\uparrow}$ & PSNR~$\textcolor{black}{\uparrow}$ & SSIM~$\textcolor{black}{\uparrow}$ \\
\midrule[0.15em]
RESCAN~\cite{li2018recurrent} & ECCV2018 & 25.09 & 0.837 & 22.55 & 0.727  \\

PReNet~\cite{ren2019progressive} & CVPR2019 & 25.61 & 0.884 & 22.99 & 0.787 \\

SPDNet~\cite{yi2021structure} & ICCV2021 & 28.82 & 0.896 & 24.89 & 0.792\\

PCNet~\cite{jiang2021rain} & TIP2021 & 25.68 & 0.837 & 22.89 & 0.726\\

MPRNet~\cite{Zamir2021MPRNet} & CVPR2021 & 29.96 & 0.916 & 25.58 & 0.815   \\

HINet~\cite{Chen_2021_CVPR} & CVPRW2021 &  25.81 & 0.882 & 23.17 & 0.787 \\

DANet~\cite{jiangDANet2022} & IJCAI2022 & 29.54 & 0.914 & 25.28 & 0.812  \\

Uformer~\cite{Wang_2022_CVPR} & CVPR2022 & 28.99 & 0.903 & 25.02 & 0.803 \\

ALformer~\cite{magic_elf} & ACMMM2022 & 29.11 & 0.911 & 25.11 & 0.809 \\

NAFNet~\cite{chen2022simple} & ECCV2022 & 29.81 & 0.907 & 25.33 & 0.806 \\

MFDNet~\cite{10348527} & TIP2023 &  28.57 & 0.882 & 24.53 & 0.766 \\

HCT-FFN~\cite{chen2023hybrid} & AAAI2023 & 28.09 & 0.891 & 24.48 & 0.791 \\

DRSformer~\cite{Chen_2023_CVPR} & CVPR2023 & 30.83 & 0.923 & 25.86 & 0.819 \\

ChaIR~\cite{cui2023exploring} & KBS2023 & 30.88 & \underline{0.925} & 25.84 & \underline{0.820}   \\

IRNeXT~\cite{cui_ir} & ICML2023 & 30.69 & 0.924 & 25.79 & 0.819 \\

OKNet~\cite{cui2024omni} & AAAI2024 & 30.39 & 0.924 & 25.65 & 0.818 \\

SFHformer~\cite{SFHformer} & ECCV2024 & 23.09 & 0.869 & 21.23 & 0.772   \\

Nerd-rain~\cite{NeRD-Rain} & CVPR2024 & 30.96 & 0.924 & 25.96 & 0.819 \\

MSDT~\cite{chen2024rethinking} & AAAI2024 & 30.85 & 0.922 & 25.89 & 0.818 \\

FSNet~\cite{FSNet} & TPAMI2024 & 30.83 & \underline{0.925} & \underline{25.99} & 0.819   \\

AdaIR~\cite{cui2025adair} & ICLR2025 & \underline{30.99} & 0.924 & 25.97 & 0.817 \\

\midrule

\textbf{CPRAformer (Ours)} & -- & \textbf{31.19} & \textbf{0.926} & \textbf{26.01} & \textbf{0.821}  \\

\bottomrule[0.15em]
\end{tabular}}

\end{center}
\vspace{-4mm}
\end{table}

\vspace{-1mm}
{\flushleft\textbf{Efficient Prompt Guide Operator.}}
Since a fixed \( k \) leads to a rigid pattern structure that cannot dynamically update with the neural network, it struggles to adapt to the complex rain conditions in real-world scenarios. To address this, we propose the EPGO, which provides the neural network with dynamic prompt information. This guides the attention matrix towards optimal selection, allowing the model to optimize attention allocation in both sparse and dense attention scenarios, preserving the most valuable information in the current features.
As shown in Fig. \ref{overview}, the architecture of EPGO works as follows: given the input feature \( F \in \mathbb{R}^{H \times W \times C} \), two linear layers with hidden ReLU activation and a Sigmoid function are first used to generate the prompt guide features. These features are then flattened into a low-dimensional vector. Finally, the low-dimensional vector is averaged and multiplied element-wise with the channel \( C \) of the input feature \( F \), adapting to the dimensional changes in the features and effectively setting a soft threshold for the \( K \) values. 

\subsection{Spatial Pixel Refinement Self-Attention}
Unlike SPC-SA, the motivation of SPR-SA is to efficiently model spatial information for each pixel and effectively extract local details. However, existing spatial self-attention mechanisms often come with high computational costs and have limited capability in modeling local details \cite{Wang_2022_CVPR}. 
To address this, we propose a simple yet efficient self-attention approximation module using convolutions. By capturing the positional information of each pixel, the features are reweighted, allowing each pixel to perceive different degradation signals from the same position across all channels. This approach enhances the model's ability to handle spatial variations in deraining tasks. 
The specific operations are illustrated in Fig. \ref{overview}, and the computation is formulated as follows: 
\vspace{-1mm}
\begin{equation}
    F_{L}=PW(DW^{3\times 3}(Linear(F))),
    \vspace{-1mm}
\end{equation}
\begin{equation}
    F_{SP}=GAP(F_{L}),
   \vspace{-1mm}
\end{equation}
\begin{equation}
    F'=PW(\varphi (F_{SP} \odot F_{L})),
    \vspace{-1mm}
\end{equation}
where \( \it{PW}(\cdot) \) denotes point-wise convolution, \( \it{DW}(\cdot)^{x \times x } \) represents \( x \times x \) depth-wise convolutions, \( \it{GAP}(\cdot) \) stands for global average pooling, and \( \varphi(\cdot) \) refers to the GELU function.

\begin{figure*}[ht]
	\centering
\includegraphics[width=\linewidth]{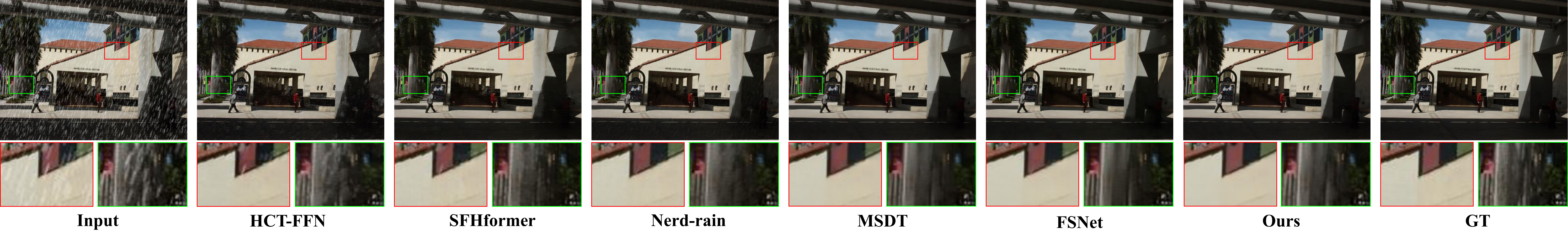} 
	\vspace{-7mm}
	\caption{The qualitative comparison on Test100 \cite{zhang2019image}. See the \textcolor{purple}{supplements} for more visualizations.}
	\label{test100}
	\vspace{-2mm}
\end{figure*}
\begin{figure*}[ht]
	\centering
	\includegraphics[width=\linewidth]{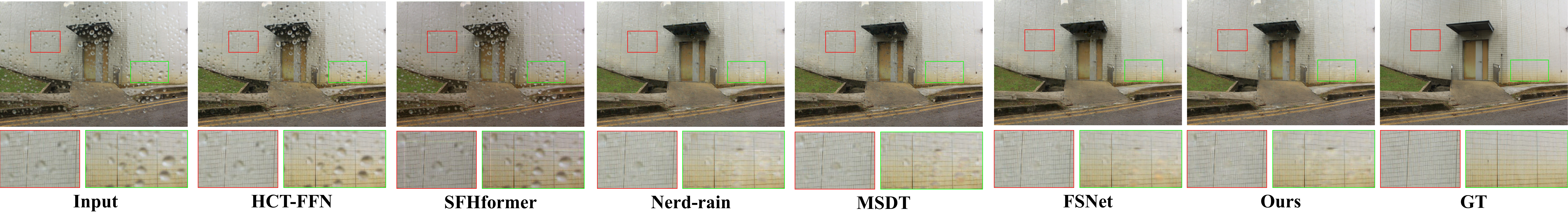} 
	\vspace{-7mm}
	\caption{The qualitative comparison on raindrop datasets \cite{Qian_2018_CVPR}. Our result has the best visual quality and details.}
	\label{raindrop}
	\vspace{-2mm}
\end{figure*}
\begin{figure*}[t]
	\centering
	\includegraphics[width=\linewidth]{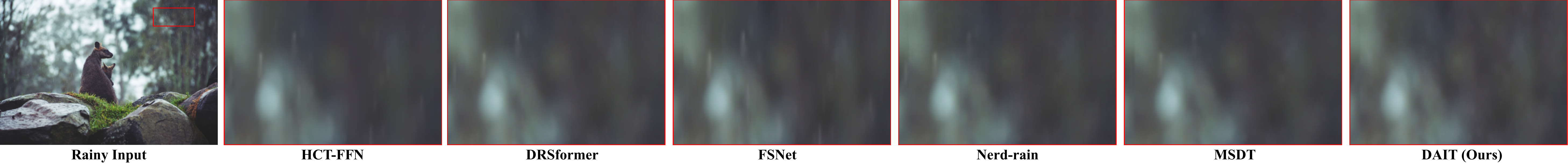} 
	\vspace{-7mm}
	\caption{Comparison of visual results on a real-world dataset \cite{wang2019spatial}. Our result has the best visual quality and details.}
	\label{real}
	\vspace{-2mm}
\end{figure*}

\subsection{Adaptive Alignment Frequency Module}
Although SPC-SA and SPR-SA capture global channel features and local spatial features respectively, effectively integrating the information from these two branches becomes a critical challenge. An intuitive observation is that there exists an uncertain knowledge gap between the convolution-based local features from CNN and the self-attention-based global features from Transformer, as well as between spatial and channel dimension features \cite{chen2023dual,chen2023hybrid}. Thus, simply concatenating or adding these features cannot fully maximize their potential.
To address this issue, we propose the AAFM. AAFM adopts a two-stage process that progressively integrates the features from both paradigms layer by layer. First, based on the types of features from the two branches, AAFM adaptively reweights the features along either the spatial or channel dimensions to align the first paradigm (i.e., spatial-channel). Then, we introduce the features into the frequency domain, leveraging the Fourier transform to enhance the interaction across multiple frequency spaces. This allows each pixel to perceive patterns from other pixels, achieving the aggregation of global features into local features and the diffusion of local features into global ones, thus realizing the deep interaction and fusion of the second paradigm (i.e., global-local).
Specifically, given the input features \( F_{spc} \) and \( F_{spr} \) from the two branches, AAFM first applies two interaction operations: Spatial Alignment and Channel Alignment, which generate spatial attention maps and channel attention maps, respectively. These are then reweighted onto the corresponding branch features to achieve effective alignment and interaction. The process can be expressed as follows:

\begin{equation}
    Map_{S}=f(PW(\varphi (PW(F_{SPC})))),
   \vspace{-1mm}
\end{equation}
\begin{equation}
    Map_{C}=f(PW(\varphi (PW(GAP(F_{SPR}))))),
   \vspace{-1mm}
\end{equation}
\begin{equation}
    \hat{F}=F_{SPR}\odot Map_{S}+ F_{SPC} \odot Map_{C}.
    \vspace{-1mm}
\end{equation}

\noindent
To leverage feature differences in the frequency domain and bridge information gaps, we apply the Fast Fourier Transform (FFT) to the fused feature \( \hat{F} \). 
For simplicity, let us first consider a single-channel case, \( \hat{F} \in \mathbb{R}^{H \times W} \). 
The 2D Fast Fourier Transform of \( \hat{F} \) is defined as:
\begin{equation}
\label{eq:2dfft}
\mathcal{F}(\hat{F})(u, v) = \frac{1}{\sqrt{HW}} \sum_{h=0}^{H-1} \sum_{w=0}^{W-1} \hat{F}(h, w)e^{-j2\pi \bigl(\frac{u h}{H} + \frac{v w}{W}\bigr)}, 
\end{equation} 
where \(u\) and \(v\) denote the frequency coordinates in the transformed space, and \(\mathcal{F}^{-1}\) denotes the corresponding inverse transform (IFFT). 
\vspace{2pt}
\noindent
In our implementation, before applying FFT, the feature \( \hat{F} \in \mathbb{R}^{H \times W \times C} \) is first projected by a linear layer to amplify high-frequency signals, acting as a high-pass filter \cite{SFHformer,park2022vision,wang2022antioversmooth}. We then use FFT to decompose the feature into real and imaginary parts, \( (R, I) \). Mathematically, we have:
\begin{equation}
    R,\, I \;=\; \text{FFT}\!\bigl(\text{Linear}(\hat{F})\bigr),
    \vspace{-1mm}
\end{equation}
which produces a complex spectrum containing crucial global information. Next, these real and imaginary components are concatenated along the channel dimension:
\begin{equation}
    F \;=\; \text{IFFT}\!\bigl(\text{Linear}([R,\, I])\bigr),
    \vspace{-1mm}
\end{equation}
where \([\cdot]\) denotes channel concatenation, and \(F\) is the frequency-domain interaction-fused feature. This second linear layer further modulates and refines the frequency components, while the inverse FFT transforms them back to the spatial domain for subsequent restoration stages.

\subsection{Multi-Scale Flow Gating Network}
\label{GMFN}
Traditional feed-forward networks often rely on deep convolutions to enhance the locality of latent features, but they tend to overlook the effectiveness of multi-scale feature representation in removing rain streaks \cite{Chen_2023_CVPR,chen2024rethinking}. To address this, we combine a gating mechanism with multi-scale representation learning by introducing convolutions of different scales into both the gating and value branches (see Fig. \ref{overview}). This controls the flow of expert information across the various levels of our pipeline, facilitating the flow of cross-level, multi-scale information and effectively extracting multi-scale local details. 
Given the input feature \( F \), the computation of MSGN can be formulated as: 
\vspace{-1mm}
\begin{equation}
    \hat{F} =PW(F) ,[\hat{F_{1}},\hat{F_{2}} ]=\hat{F}, 
    \vspace{-1mm}
\end{equation}
\begin{equation}
    \hat{F}'=Linear(DW^{3\times 3}(\hat{F_{1}} )\odot DW^{5\times 5}(\hat{F_{2}} )). 
\end{equation}

\section{Experiments}
\subsection{Experimental Settings}
{\flushleft\textbf{Implementation Details.}}
In CPRAformer, \( \{N_1, N_2, N_3, N_4\} \) are set to \{4, 6, 6, 8\}, and the attention heads of the four levels of CPRAformerB are set to \{1, 2, 4, 8\}. The initial channel \( C \) is set to 48. During training, we use the Adam optimizer with a patch size of 64×64, a batch size of 12, and the number of epochs set to 300. For detailed experimental settings, please refer to the \textcolor{purple}{supplements}.
{\flushleft\textbf{Data and Evaluation.}}
We follow the majority of previous research practices to train and validate our model \cite{Zamir2021Restormer,Zamir2021MPRNet,FSNet,Chen_2021_CVPR,magic_elf}. Specifically, we use 13,712 image pairs collected from multiple datasets \cite{8099669,7780668,yang2017deep,derain_zhang_2018,zhang2019image} for training and evaluate the model on five synthetic datasets (Test100 \cite{zhang2019image}, Rain100H \cite{yang2017deep}, Rain100L \cite{yang2017deep}, Test2800 \cite{fu2017removing}, Test1200 \cite{zhang2018density}) and one real-world dataset \cite{wang2019spatial}. The aforementioned datasets primarily target rain streak removal. However, raindrops represent another form of contamination in rain removal tasks. Therefore, we train and evaluate our model on the Raindrop-A and Raindrop-B datasets \cite{Qian_2018_CVPR}. Consistent with existing methods \cite{10035447,Kui_2020_CVPR}, we adopt PSNR and SSIM as evaluation metrics for the aforementioned benchmarks, both calculated on the Y channel (luminance) in the YCbCr color space.

\subsection{Comparison with State-of-the-Art Methods}
We compared CPRAformer with 22 image deraining methods: RESCAN \cite{li2018recurrent},
PreNet \cite{ren2019progressive}, SPDNet \cite{yi2021structure}, PCNet \cite{jiang2021rain}, MPRNet \cite{Zamir2021MPRNet}, HINet \cite{Chen_2021_CVPR}, ALformer \cite{magic_elf}, DANet \cite{jiangDANet2022}, Uformer \cite{Wang_2022_CVPR}, NAFNet \cite{chen2022simple}, MFDNet\cite{10348527}, HCT-FFN \cite{chen2023hybrid}, DRSformer \cite{Chen_2023_CVPR}, FSNet \cite{FSNet}, ChaIR \cite{cui2023exploring}, OKNet \cite{cui2024omni}, AST \cite{zhou2024AST}, SFHformer \cite{SFHformer}, IRNeXT \cite{cui_ir}, Nerd-rain \cite{NeRD-Rain}, MSDT \cite{chen2024rethinking} and AdaIR \cite{cui2025adair}.
To ensure a fair comparison, we retrained all the above methods from scratch in our environment using their official source codes without any pretraining or fine-tuning.

\begin{figure}[t]
	\centering
	\includegraphics[width=0.85\linewidth]{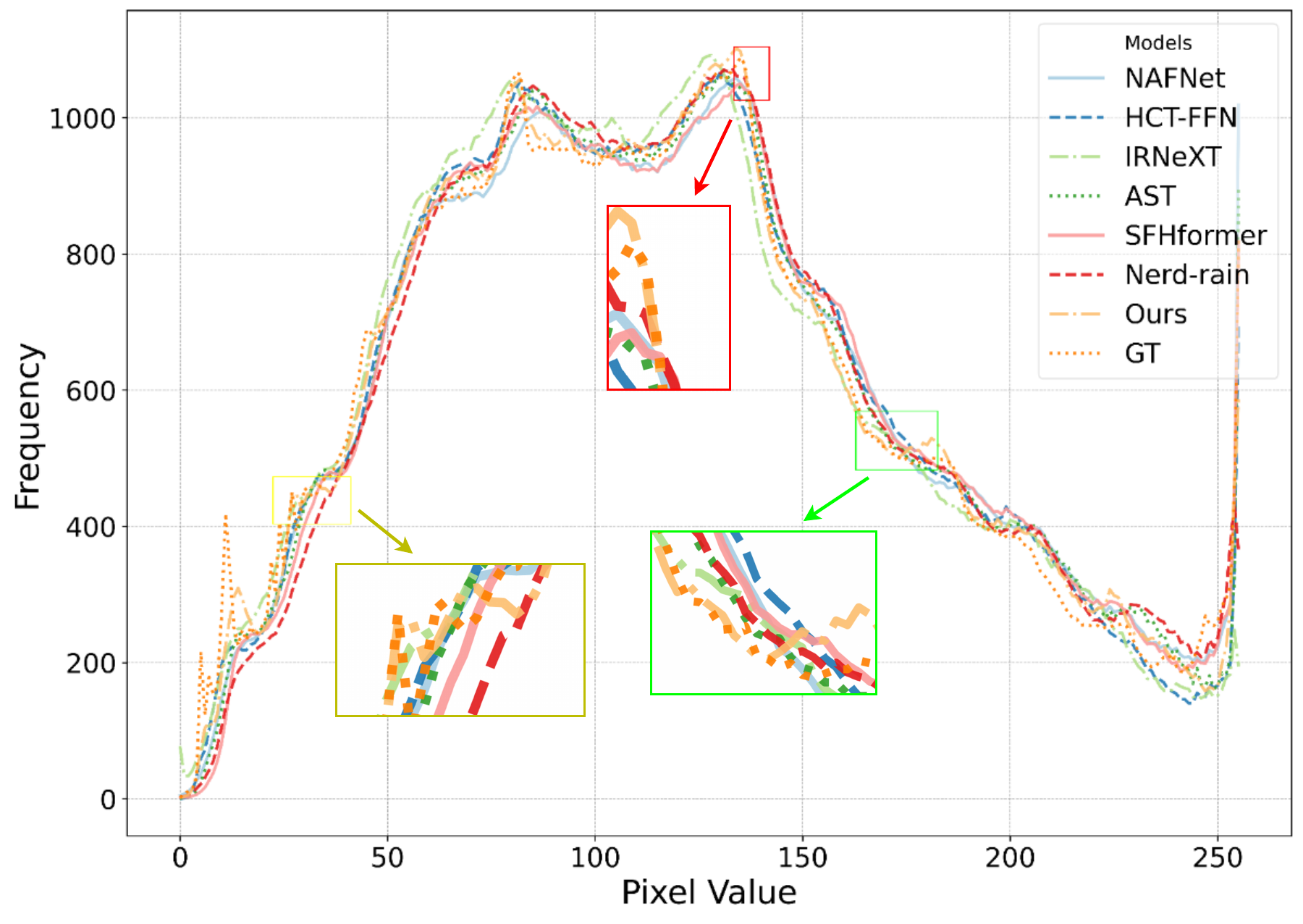} 
	\vspace{-5mm}
	\caption{The average fitting results of the Y channel histogram curve in the YCbCr space on the synthetic dataset, our method produces results most similar to the ground truth.}
	\label{y_vis}
	\vspace{-2mm}
\end{figure}

\begin{table}[t]
\begin{center}
\caption{NIQE results under the real-world scenario \cite{wang2019spatial}.}
\label{table:niqe}
\vspace{-3mm}
\setlength{\tabcolsep}{9pt}
\resizebox{1\linewidth}{!}{
\begin{tabular}{cccccccccccccccc}
\toprule[0.15em]
\multicolumn{2}{l|}{{Methods}} & \multicolumn{2}{c}{Input}   & \multicolumn{2}{c}{DRSformer \cite{Chen_2023_CVPR}} & \multicolumn{2}{c}{SFHformer}\cite{SFHformer} & \multicolumn{2}{c}{Nerd-rain \cite{NeRD-Rain}} &\multicolumn{2}{c}{FSNet \cite{FSNet}}&\multicolumn{2}{|c}{CPRAformer} \\ 
\midrule[0.1em]
\multicolumn{2}{l|}{NIQE $\downarrow$} &\multicolumn{2}{c}{5.923}   & \multicolumn{2}{c}{5.814}  & \multicolumn{2}{c}{5.745} & \multicolumn{2}{c}{5.711} & \multicolumn{2}{c}{5.667}& \multicolumn{2}{|c}{\textbf{5.556}}    \\
\bottomrule[0.15em]
\end{tabular}
}
\end{center}
\vspace{-3mm}
\end{table}

\begin{table}[t]
\begin{center}
\caption{Ablation study of dual aggregation and dual paradigm strategy.}
\label{table:ablation1}
\vspace{-3mm}
\setlength{\tabcolsep}{9pt}
\resizebox{1\linewidth}{!}{
\begin{tabular}{l c c  c c c || c c}
\toprule[0.15em]
  & & \multicolumn{2}{c}{\textbf{Test100}}&\multicolumn{2}{c||}{\textbf{Rain100H}}&\multicolumn{2}{c}{\textbf{Average}}\\
 \textbf{SPC-SA} & \textbf{SPR-SA} &
 PSNR~$\textcolor{black}{\uparrow}$ & SSIM~$\textcolor{black}{\uparrow}$ & PSNR~$\textcolor{black}{\uparrow}$ & SSIM~$\textcolor{black}{\uparrow}$ & PSNR~$\textcolor{black}{\uparrow}$ & SSIM~$\textcolor{black}{\uparrow}$\\
\midrule[0.1em]
\CheckmarkBold &  & 27.62  & 0.868  & 28.90  & 0.866  & 30.66 & 0.903    \\

 & \CheckmarkBold & 27.40  & 0.866    & 29.01  & 0.864  & 30.91 & 0.900    \\
 \CheckmarkBold &  \CheckmarkBold & \textbf{28.80}  & \textbf{0.891}  & \textbf{29.22}  & \textbf{0.871}  & \textbf{31.56} & \textbf{0.913}    \\

\bottomrule[0.15em]
\end{tabular}}

\end{center}
\vspace{-6mm}
\end{table}

\vspace{-1mm}
{\flushleft\textbf{Rain Streak Synthetic Datasets.}}
Table \ref{table:deraining} presents the quantitative evaluation results on five benchmark datasets, where it is evident that our CPRAformer consistently and significantly outperforms existing methods. Specifically, compared to the most recent top-performing method, AdaIR \cite{cui2025adair}, CPRAformer improves the average performance across all datasets by 0.52dB. On certain datasets (such as Test100), the gain reaches up to 1.01dB. Fig. \ref{test100} shows a comparison of the visual quality of samples generated by recent methods. Thanks to the interaction and fusion of dual paradigms, CPRAformer effectively removes rain streaks while preserving details and realistic textures in the background image. Additionally, we provide a comparison of the “Y" channel histogram fitting curves in Fig. \ref{y_vis}, confirming the consistency of the deraining results with the ground truth in statistical distribution.

{\flushleft\textbf{Raindrops Synthetic Datasets.}}
We conducted further experiments on the raindrop datasets \cite{Qian_2018_CVPR}, and the quantitative results are shown in Table \ref{table:raindrop}, where our model achieved the highest performance. Visual comparisons in Fig \ref{raindrop} indicate that our method effectively removes raindrops while preserving fine texture details. This demonstrates that our model attains optimal performance under both types of rain contamination, further confirming the generalization capability of our CPRAformer.

\vspace{-1mm}
{\flushleft\textbf{Real-world Datasets.}}
To further demonstrate the generalization and robustness of CPRAformer, we conducted comparisons on real-world datasets \cite{wang2019spatial}. As shown in Fig. \ref{real}, all other methods produced suboptimal results in either rain removal or detail restoration.
\begin{figure}[t]
	\centering
\includegraphics[width=1\linewidth]{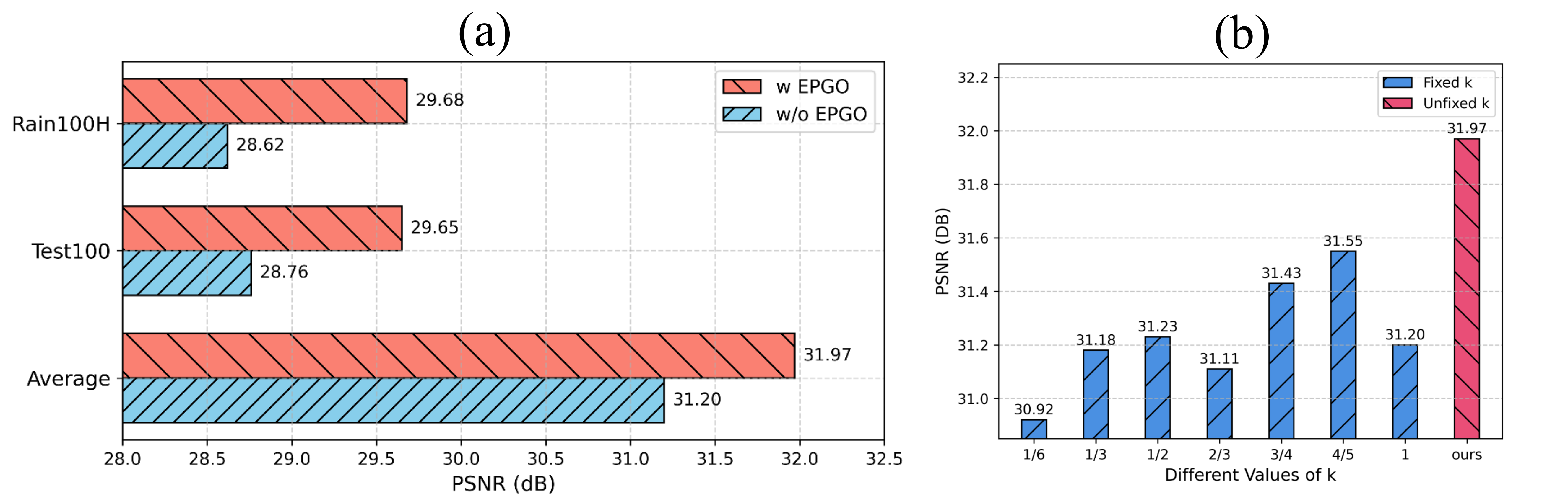} 
	\vspace{-5mm}
	\caption{Ablation study of EPGO.}
	\label{EPGO}
	\vspace{-3mm}
\end{figure}
\begin{table}[!t]
\begin{center}
\caption{Ablation study of AAFM.}
\label{table:ablation2}
\vspace{-4mm}
\setlength{\tabcolsep}{9pt}
\resizebox{1\linewidth}{!}{
\begin{tabular}{l c c c c || c c}
\toprule[0.15em]
  & \multicolumn{2}{c}{\textbf{Test100}}&\multicolumn{2}{c||}{\textbf{Rain100H}}&\multicolumn{2}{c}{\textbf{Average}}\\
 \textbf{Method}  &
 PSNR~$\textcolor{black}{\uparrow}$ & SSIM~$\textcolor{black}{\uparrow}$ & PSNR~$\textcolor{black}{\uparrow}$ & SSIM~$\textcolor{black}{\uparrow}$ & PSNR~$\textcolor{black}{\uparrow}$ & SSIM~$\textcolor{black}{\uparrow}$\\
\midrule[0.1em]
baseline & 28.80  & 0.891  & 29.22  & 0.871  & 31.56  & 0.913     \\
 w/ stage1 & 28.78 & 0.892  & 29.68    & 0.873  & 31.68  & 0.913   \\
 w/ stage2 & \textbf{29.65} & \textbf{0.895}  & \textbf{29.68}  & \textbf{0.875}  & \textbf{31.97}  & \textbf{0.916}    \\
\bottomrule[0.15em]
\end{tabular}}
\end{center}
\vspace{-3mm}
\end{table}
\begin{table}[!t]
\begin{center}
\caption{Ablation study of MSGN.}
\label{table:ablation4}
\vspace{-4mm}
\setlength{\tabcolsep}{9pt}
\resizebox{1\linewidth}{!}{
\begin{tabular}{l c c c c || c c}
\toprule[0.15em]
  & \multicolumn{2}{c}{\textbf{Test100}}&\multicolumn{2}{c||}{\textbf{Rain100H}}&\multicolumn{2}{c}{\textbf{Average}}\\
 \textbf{Method}  &
 PSNR~$\textcolor{black}{\uparrow}$ & SSIM~$\textcolor{black}{\uparrow}$ & PSNR~$\textcolor{black}{\uparrow}$ & SSIM~$\textcolor{black}{\uparrow}$ & PSNR~$\textcolor{black}{\uparrow}$ & SSIM~$\textcolor{black}{\uparrow}$\\
\midrule[0.1em]
SFFN \cite{dosovitskiy2020vit}& 27.20  & 0.864  & 28.51  & 0.862  & 31.01  & 0.905     \\
DFN \cite{li2021localvit} & 29.45 & 0.893  & 29.66    & 0.872  & 31.58  & 0.913   \\
ConvGLU \cite{shi2024transnext} & 28.33 & 0.891  & 29.26  & 0.870  & 31.46  & 0.911    \\
MSGN & \textbf{29.65} & \textbf{0.895}  & \textbf{29.68}  & \textbf{0.875}  & \textbf{31.97}  & \textbf{0.916}    \\
\bottomrule[0.15em]
\end{tabular}}
\end{center}
\vspace{-4mm}
\end{table}
In contrast, our proposed CPRAformer outperformed the other methods, achieving visually pleasing restoration in challenging examples. This indicates that CPRAformer can generalize well to unseen real-world data types.

{\flushleft\textbf{Perceptual quality assessment.}}
We followed the method in \cite{Chen_2023_CVPR,zhou2024AST} to evaluate the perceptual quality of our proposed CPRAformer. The results, shown in Table \ref{table:niqe}, demonstrate that CPRAformer achieves a lower NIQE compared to other methods, indicating it delivers better perceptual quality in real rain scenes.


\subsection{Ablation Studies}
In this section, we conduct ablation experiments on five synthetic datasets to investigate the effect of each component. To ensure fair comparison, all ablation studies are performed under the same environment and training details. Due to space limits, we present results for two datasets along with the average across all five datasets. Further ablation experiments are provided in the \textcolor{purple}{supplements}.
\vspace{-2mm}
\begin{figure*}[h]
	\centering
	\includegraphics[width=\linewidth]{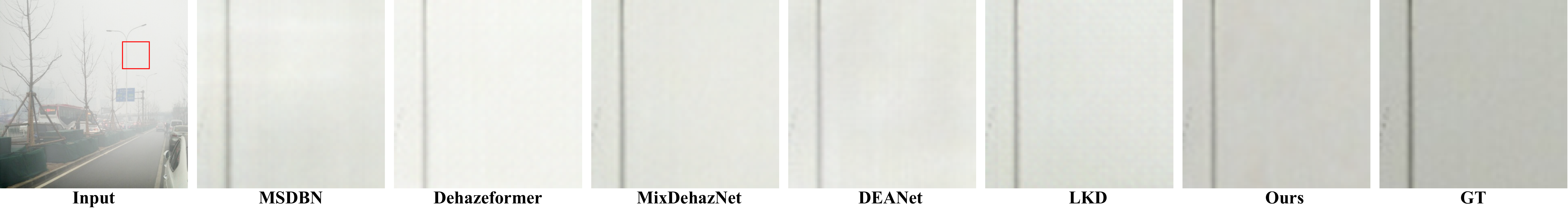} 
	\vspace{-8mm}
	\caption{The qualitative comparison on hazy images. Our result has the best visual quality and details.}
	\label{hazy}
	\vspace{-2mm}
\end{figure*}
\begin{table}
\begin{center}
\caption{Quantitative comparisons of state-of-the-art methods on the REIDE-6K \cite{li2018benchmarking,song2023vision} and Haze4K \cite{liu2021synthetic} datasets. Bold and underlined indicate the best and second-best results.}
\label{table:dehazing}
\vspace{-2mm}
\setlength{\tabcolsep}{8pt}
\scalebox{0.72}{
\begin{tabular}{l c c c c c}
\toprule[0.15em]
  & & \multicolumn{2}{c}{\textbf{RESIDE-6K}~\cite{li2018benchmarking}} & \multicolumn{2}{c}{\textbf{Haze4K}~\cite{liu2021synthetic}}\\
 \textbf{Method} & \textbf{Year} & PSNR~$\textcolor{black}{\uparrow}$ & SSIM~$\textcolor{black}{\uparrow}$ & PSNR~$\textcolor{black}{\uparrow}$ & SSIM~$\textcolor{black}{\uparrow}$\\
\midrule[0.15em]

MSBDN~\cite{MSBDN-DFF} & CVPR2020 & 25.07  & 0.896  & 25.74  & 0.918  \\

FFA-Net~\cite{qin2020ffa} & AAAI2020 & 24.44  & 0.925  & 27.38  & 0.942   \\

UHD~\cite{xiao2022singleuhdimagedehazing} & CVPR2021 & 25.68  & 0.913  & 25.40  & 0.918   \\

Uformer~\cite{Wang_2022_CVPR} & CVPR2022 & 26.29  & 0.925  & 26.43  & 0.937   \\

gUNet~\cite{song2022vision} & Arxiv2022 & 25.67  & 0.924  & 26.50  & 0.940   \\

LKD~\cite{10219934} & ICME2023 & 25.42  & 0.925  & 27.39  & 0.938   \\

Dehazeformer~\cite{song2023vision} & TIP2023 & 26.25  & 0.931  & \underline{27.45}  & \underline{0.946}   \\

MB-TaylorFormer~\cite{2308.14036} & ICCV2023 & 26.28  & 0.923  & 26.34  & 0.933   \\

MixDehazeNet ~\cite{lu2024mixdehazenet} & IJCNN2024 & 26.62  & 0.939  & 27.34  & 0.945   \\

DEANet ~\cite{chen2023dea} & TIP2024 & 26.61  & 0.932  & 26.94  & 0.942   \\

SFHformer ~\cite{SFHformer} & ECCV2024 & \underline{27.08}  & \underline{0.940}  & 26.92  & 0.941   \\

\midrule

\textbf{CPRAformer (Ours)} & -- & \textbf{27.70} & \textbf{0.944} & \textbf{27.97} & \textbf{0.952}  \\

\bottomrule[0.1em]
\end{tabular}}

\end{center}
\vspace{-4mm}
\end{table}


\vspace{2mm}
{\flushleft\textbf{Dual Aggregation and Dual Paradigm Strategy.}}
To investigate the effect of using both SPC-SA and SPR-SA simultaneously, we conducted multiple experiments, with results shown in Table \ref{table:ablation1}. The first and second rows of the table indicate that we replaced all attention modules in CPRAformer with either SPC-SA or SPR-SA. The third row represents the scenario where both attention mechanisms are used in CPRAformer. Additionally, none of the models employed AAFM in this comparison. We observe that the best performance of 31.56 dB is achieved when both types of self-attention are utilized. This indicates that the different representations of the two paradigms are crucial for high-quality image deraining.

\vspace{-1mm}
{\flushleft\textbf{Efficient Prompt Guide Operator.}}
To verify the effectiveness of EPGO, we first removed the Top-k mechanism and EPGO itself, with experimental results shown in Fig. \ref{EPGO} \textcolor{red}{(a)}. It is observed that EPGO consistently achieves high-fidelity recovery with excellent PSNR performance. The key aspect of EPGO is to generate dynamic k values through prompt information, guiding the attention matrix to filter out important information. To this end, we tested on five datasets, comparing the dynamic k values generated by EPGO with fixed k values, with average results shown in Fig. \ref{EPGO} \textcolor{red}{(b)}. The dynamic k values adaptively handle the complex and variable rain streaks in real scenarios, enhancing the model's robustness.

\vspace{-1mm}
{\flushleft\textbf{Adaptive Alignment Frequency Module.}}
We validated the effectiveness of the AAFM through comprehensive ablation experiments, with results shown in Table \ref{table:ablation2}. Specifically, we used the model from the third row of Table \ref{table:ablation1} as the baseline. First, we introduced the first alignment stage of AAFM, which achieved a gain of 0.12 dB. Subsequently, we added the second fusion stage, resulting in a gain of 0.41 dB compared to the baseline model. This indicates that AAFM maximizes the feature advantages of both paradigms, facilitating high-frequency interactions and deep fusion of information.
\vspace{-1mm}

{\flushleft\textbf{Multi-Scale Flow Gating Network.}}
To validate the effectiveness of MSGN, we replaced it with Standard Feed-Forward Network (SFFN) \cite{dosovitskiy2020vit}, Depth-wise Convolution Equipped Feed-Forward Network (DFN) \cite{li2021localvit}, and Convolutional Gated Linear Unit (ConvGLU) \cite{shi2024transnext}. The results in Table \ref{table:ablation4} indicate that MSGN achieves optimal performance by effectively representing latent multi-scale perceptual information and enhancing information flow between classes.

\subsection{Other Related Tasks}
We selected image dehazing to validate the extensibility and robustness of CPRAformer in image restoration tasks. Following most previous studies \cite{chen2023dea,lu2024mixdehazenet,song2023vision,song2022vision}, we trained and validated our model. Specifically, we used two popular dehazing datasets, RESIDE-6K \cite{li2018benchmarking,song2023vision} and Haze-4K \cite{liu2021synthetic}, and compared CPRAformer with 11 state-of-the-art dehazing methods. 
\begin{figure}[t]
	\centering
	\includegraphics[width=\linewidth]{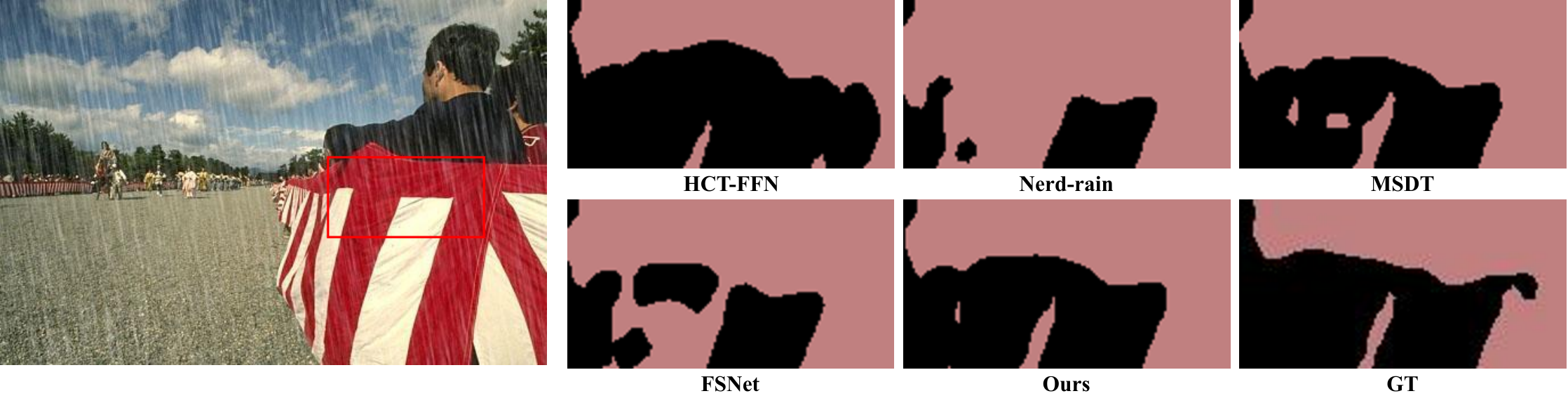} 
	\vspace{-8mm}
	\caption{Semantic segmentation results on Deeplab V3 \cite{deeplabv3plus2018}.}
	\label{seg}
	\vspace{-3mm}
\end{figure}
Quantitative results, as shown in Table \ref{table:dehazing}, indicate that CPRAformer achieved the best performance on both datasets. For example, on RESIDE-6K, CPRAformer improved PSNR by 0.62 dB over the previous SOTA method SFHformer, which is a significant enhancement. Furthermore, visual comparisons in Fig. \ref{hazy} reveal that other methods produce images with unnatural shadows and high-frequency regions, as well as residual haze. In contrast, CPRAformer restores clear images, preserves texture and color details, and minimizes haze remnants. 
Additionally, we evaluated CPRAformer on downstream tasks. 
For semantic segmentation, a pre-trained DeepLab v3 \cite{deeplabv3plus2018} was used, and as shown in Fig. \ref{seg}, CPRAformer's output is closest to the ground truth. Further details on the downstream tasks are provided in the \textcolor{purple}{supplements}.



\vspace{-1mm}
\section{Conclusion}
In this paper, we propose a novel image deraining Transformer model, CPRAformer, based on dual-paradigm representation learning. Specifically, we introduce SPC-SA, which adaptively adjusts the sparsity of the neural network, enhancing global modeling capability while facilitating the flow of expert information across channels. Additionally, SPR-SA emphasizes the spatial distribution of rain variations, focusing on local feature extraction. Furthermore, we propose AAFM and MSGN to fully integrate features from both paradigms, promoting interaction between different types of representations and achieving cross-scale feature interaction. Extensive experiments on 10 benchmark datasets demonstrate that CPRAformer exhibits strong generalization and robustness.
\begin{acks}
This work was supported by the Open Fund Project of Provincial Key Laboratory for Computer Information Processing Technology (Soochow University) under Grant KJS2274.
\end{acks}

\bibliographystyle{ACM-Reference-Format}
\bibliography{sample-base}

\end{document}